\documentclass{article}

\PassOptionsToPackage{numbers, compress}{natbib}
\usepackage[preprint]{neurips_2026}

\usepackage[utf8]{inputenc} 
\usepackage[T1]{fontenc}    
\usepackage{hyperref}       
\usepackage{url}            
\usepackage{booktabs}       
\usepackage{amsfonts}       
\usepackage{nicefrac}       
\usepackage{microtype}      
\usepackage{xcolor}         

\definecolor{myframepink}{RGB}{72,138,176}

\hypersetup{
    colorlinks=true,
    allcolors=myframepink
}

\usepackage{amsmath}
\usepackage{tabularx}
\usepackage{arydshln}
\usepackage{graphicx}
\usepackage{longtable,array,calc}

\title{When Does AI for PDEs Yield Scientific Evidence?}

\author{
Wenshuo Wang \\
School of Future Technology, South China University of Technology, China \\
\texttt{202364870251@mail.scut.edu.cn}
}

\begin{document}

\maketitle

\begingroup
\renewcommand{\thefootnote}{}
\footnotetext{Our code is available at
\url{https://github.com/VinceWangVW/ai-pde-scientific-evidence}.}
\endgroup

\begin{abstract}
  Existing AI-for-PDE benchmarks primarily assess models in terms of predictive or approximation accuracy. In physics research, however, AI outputs often serve as evidence for scientific claims. These two objectives are not equivalent: the former measures an output's agreement with a reference target or satisfaction of governing constraints; the latter asks whether, given a specified object of study, scientific claim, assumptions, and evidence standard, the output provides sufficient evidence for that claim. To bridge this gap, we extend a widely used PDE-simulation benchmark and a comprehensive benchmark for PDE inverse problems to enable, for the first time in AI for PDEs, evaluation of whether and to what extent model outputs support specified scientific claims. Our results show that numerical accuracy and evidential support can rank models differently, explain when and why they do so, and reveal that existing benchmarks can favor methods whose outputs provide weaker support for the scientific claims of interest. Together, we formalize, empirically demonstrate, and explain this evaluation--use mismatch in AI for PDEs.
\end{abstract}

\section{Introduction}

As reviewed by~\citet{karniadakis2021physics}, representative tasks in AI for PDEs fall into two broad families: forward PDE solving and simulation~\citep{takamoto2022pdebench,hao2024pinnacle} and PDE-constrained inverse problems~\citep{zhao2022inverse}. Despite their different formulations, these benchmarks predominantly evaluate how closely model outputs agree with reference objects or governing constraints---that is, their predictive or approximation accuracy---as summarized in Section~\ref{sec:benchmark-use}. In scientific research, however, the same outputs often serve as evidence for scientific claims. For example, in forward PDE solving, PINN-derived self-similar profiles provide numerical evidence for candidate singularity scenarios in the Boussinesq and Euler equations~\citep{wang2023blowup}; in an inverse problem, rheological parameters and viscosity fields inferred from remote-sensing data support claims about Antarctic ice-shelf rheology~\citep{wang2025ice}.

We formalize and distinguish the numerical accuracy evaluated by AI-for-PDE benchmarks from the scientific evidence required to support scientific claims:
\[
q(A;R,K) \qquad \text{vs.} \qquad W(A;O,C,H,E).
\]
Here, \(q\) evaluates how closely a model output \(A\) agrees with a reference object \(R\) or governing constraints \(K\), whereas \(W\) evaluates whether \(A\) provides sufficient evidence for a claim \(C\) about an object of study \(O\), under assumptions \(H\) and evidence standard \(E\). As formalized in Section~\ref{sec:mismatch}, better \(q\) does not generally imply that \(W=1\). This distinction is fundamental because benchmarks shape which methods the community selects and improves: optimizing \(q\) can favor methods whose outputs provide weaker support for the scientific claims for which they are ultimately used.

To evaluate this distinction, we extend selected forward-simulation tasks from the widely used PDEBench~\citep{takamoto2022pdebench,hao2024dpot,hagnberger2024vcnef,chen2025omniarch} and inverse tasks from PDEInvBench~\citep{goel2026pdeinvbench}. We retain the original equations, datasets, task definitions, and numerical-accuracy metrics, while adding prespecified scientific claims and claim-specific evidence evaluations: quantity-of-interest and event claims for predicted solution fields, and range, regime, and regional-structure claims for inferred physical parameters or fields. This yields paired accuracy- and evidence-based evaluations on otherwise unchanged tasks, enabling direct comparison of the method selections induced by the two objectives, as detailed in Section~\ref{sec:benchmark-extensions}.

With these paired evaluations in place, our experiments establish three findings. 
\textcircled{1} The evaluation attains near-nominal coverage, gives informative verdicts, and responds to the submitted artifact. 
\textcircled{2} Numerical accuracy and claim-conditioned support yield different method selections in both benchmark extensions: the methods selected by the two objectives differ in all three headline comparisons, and this difference persists across most tested variants. 
\textcircled{3} Controlled analyses trace these differences to how close the relevant quantity lies to the claim boundary, where and in which direction model errors occur, and how tightly the available evidence constrains that quantity, explaining when and why selections based on the two objectives align or diverge, as detailed in Section~\ref{sec:different-rankings}.

Taken together, we identify and formalize an evaluation--use mismatch in AI for PDEs, develop two benchmark extensions that make it directly measurable, empirically demonstrate its consequences for method selection, and explain when and why those consequences arise. These conclusions remain conditional on the studied benchmarks and prespecified claims, assumptions, and evidence standards; within that scope, they establish the need to evaluate claim-conditioned support directly rather than infer it from numerical accuracy alone, as discussed in Section~\ref{sec:discussion}.

\section{Evaluation in AI-for-PDE Benchmarks and Use of Model Outputs as Scientific Evidence}
\label{sec:benchmark-use}

We first distinguish two bodies of AI-for-PDE research by principal contribution rather than publication venue.\footnote{Because the two bodies of work are defined by their principal contributions independently of the evaluation criteria analyzed below, their definitions do not presuppose a difference between their evaluation targets.} AI-for-PDE evaluation research comprises studies centered on reusable benchmarks or evaluation protocols and empirical evaluation components of papers on new AI-for-PDE methods or tools; the latter may reuse established protocols or introduce new ones. AI-enabled scientific applications comprise studies centered on a scientific conclusion about a specified physical system or mathematical object, with AI-generated PDE-related outputs integral to the supporting analysis.

\subsection{Evaluation in AI-for-PDE Benchmarks}
\label{sec:ai-pde-evaluation}

Following the broad forward--inverse organization of physics-informed machine learning~\citep{karniadakis2021physics}, Table~\ref{tab:ai-pde-evaluation} identifies each task and its source studies, then records the elements needed to specify how that task is evaluated: the output scored, the reference target, any constraint tested at evaluation time, and the reported criterion. Within these families, F1--F4 and I1--I6 are our operational synthesis of evaluated outputs and input--output relations; F4 denotes a cross-cutting distribution-valued forward mode, whereas inverse posteriors are grouped by the inferred quantity.

\begin{table*}[t]
  \centering
  \caption{Representative AI-for-PDE evaluation tasks. Only test-stage references, constraints, and criteria are reported; training-only losses are excluded.}
  \label{tab:ai-pde-evaluation}
  \footnotesize
  \setlength{\tabcolsep}{2pt}
  \renewcommand{\arraystretch}{1.08}
  \begin{tabularx}{\textwidth}{@{}>{\raggedright\arraybackslash}p{2.25cm}>{\raggedright\arraybackslash}p{1.15cm}>{\raggedright\arraybackslash}p{1.95cm}>{\raggedright\arraybackslash}p{2.00cm}>{\raggedright\arraybackslash}p{2.10cm}>{\raggedright\arraybackslash}X@{}}
    \toprule
    Task & Studies & Evaluated output & Reference target & Tested constraint & Reported criterion \\
    \midrule
    \multicolumn{6}{@{}l}{\textbf{Forward}} \\
    \cdashline{1-6}[1.0pt/1.2pt]
    \addlinespace[1pt]
    F1: Single-instance solution approximation & \citep{hao2024pinnacle,datar2026frozen} & Solution field & Analytic or high-fidelity solution & --- & Solution-field error \\
    F2: Parametric operator learning & \citep{huang2026domain,bischof2025hypino} & Solution for an unseen instance or geometry & Numerical or manufactured solution & --- & Solution-field error \\
    F3: Time stepping and rollout & \citep{chen2025omniarch,ohana2024well,hou2026cfo} & Future trajectory & High-fidelity simulated trajectory & --- & One-step or rollout error \\
    F4: Probabilistic forward prediction & \citep{bulte2025pno,magnani2025luno} & Predictive distribution & Held-out field or trajectory realizations & --- & Probabilistic score, calibration or coverage, and predictive-mean error \\
    \midrule
    \multicolumn{6}{@{}l}{\textbf{Inverse}} \\
    \cdashline{1-6}[1.0pt/1.2pt]
    \addlinespace[1pt]
    I1: Hidden-state or full-field reconstruction & \citep{zhou2026incomplete} & Completed field or trajectory & Complete field or trajectory, where available & Governing PDE or forward dynamics, dataset-specific & Reconstruction, residual, or propagation error \\
    I2: Initial-state recovery & \citep{huang2024diffusionpde} & Initial state & Ground-truth initial state & --- & Initial-state error \\
    I3: Parameter or coefficient inference & \citep{goel2026pdeinvbench,moss2025fnope} & Parameter, coefficient field, or posterior & True parameter or coefficient & Governing PDE or forward dynamics, where tested & Parameter or field error; posterior or consistency diagnostics \\
    I4: Source or forcing inference & \citep{cho2025pidion} & Source function & Ground-truth source & --- & Source-function error \\
    I5: Geometry or boundary-shape reconstruction & \citep{gao2022inverse} & Boundary curve or shape & Target curve & --- & Curve error \\
    I6: Governing-equation discovery & \citep{ziaei2026mdbench,gao2025ablpde} & Symbolic structure and coefficients & True equation and coefficients; held-out derivatives & --- & Structural success, coefficient or derivative error, and complexity \\
    \bottomrule
  \end{tabularx}
\end{table*}

Forward benchmarks favor methods that best reproduce reference fields or trajectories in F1--F3 and perform best under the reported distributional criteria in F4. Across I1--I6, inverse benchmarks favor methods whose inferred quantities or posterior distributions, as applicable, best agree with withheld references, satisfy tested dynamical constraints, meet uncertainty criteria, or recover the reference structure parsimoniously. Thus, despite heterogeneous metrics, both families predominantly assess the predictive or approximation quality of model outputs through agreement with references or observations, satisfaction of tested constraints, or prespecified probabilistic or structural criteria.

\subsection{Use of Model Outputs as Scientific Evidence}
\label{sec:scientific-evidence-use}

Two applications---one per broad task family---make explicit the scientific claim, linking assumptions, and checks applied to an AI-generated PDE output.

\paragraph{Forward solving and simulation.}
\citet{wang2023blowup} ask whether a smooth self-similar numerical candidate in a reduced two-dimensional Boussinesq system is consistent with the Luo--Hou axisymmetric Euler singularity scenario that the system asymptotically represents. PINN-computed self-similar fields and a scaling exponent serve as evidence under that representation and ansatz, prescribed symmetry, normalization, and far-field conditions, and an adequate finite-domain approximation. Equation residuals, robustness to domain size, initialization, and normalization, exponent agreement with simulations, and tests on known solutions support consistency with the scenario, not proof of finite-time Euler blow-up.

\paragraph{Inverse inference.}
\citet{wang2025ice} study compression zones in five Antarctic ice shelves and assess the claim that most exhibit grain-size-sensitive composite rheology spanning grain-boundary sliding, mixed, and dislocation-creep regimes, with Amery as an explicit exception. PINN-inferred viscosity fields, together with resulting strain-rate, stress, and stress-exponent estimates derived from remote-sensing observations, serve as evidence under steady two-dimensional shallow-shelf dynamics and the transferability of laboratory rheology to the resolved scale. Repeated fits and random initializations, comparisons across shelves and flow lines, and agreement with laboratory flow-law exponents and their expected dependence on stress, temperature, and grain size support this interpretation for most studied shelves.

Across both applications, support for a scoped claim about a specified physical or mathematical object is evaluated by treating an AI-generated PDE output as evidence, interpreting that output under assumptions connecting it to the object, and applying claim-relevant checks.

\section{The Evaluation--Use Mismatch}
\label{sec:mismatch}

Section~\ref{sec:benchmark-use} shows that AI-for-PDE evaluations score model outputs against reference targets or tested constraints, whereas scientific applications use those outputs to support scoped claims under linking assumptions and claim-relevant checks. We formalize these practices as two evaluation targets:
\begin{equation}
q(A;R,K),
\qquad
W(A;O,C,H,E).
\label{eq:evaluation-targets}
\end{equation}
Here, \(q\) is an oriented scalar per-instance criterion measuring agreement with reference \(R\) and/or satisfaction of tested constraints \(K\), with additional criteria kept separate. By contrast, \(W\) asks whether \(A\), together with its accompanying evaluation record and checks, provides sufficient evidence for a scoped claim \(C\) about an object of study \(O\), under assumptions \(H\) and an evidence standard \(E\). We set \(W=1\) exactly when, under \(H\), this evidential material meets the acceptance criteria that \(E\) prespecifies for \(C\). Across prespecified standards, those for which \(W=1\) characterize support without creating a claim-independent evidence score. The distinction is therefore not between a continuous numerical metric and a binary verdict, but between evaluation functions with different input signatures, evaluated objects, and semantics; absent a claim-specific bridge, one does not in general determine the other.\footnote{For a claim \(C\colon J(u)\geq\tau\), suppose \(\lvert J(u)-J(A)\rvert\leq\varepsilon(A)\) is a sound bound under \(H\) and the resulting certificate satisfies \(E\). If \(J(A)-\varepsilon(A)\geq\tau\), then \(W(A;O,C,H,E)=1\). In this restricted case, support reduces to a thresholded certified-accuracy test; without such a claim-specific bridge, numerical accuracy alone does not determine \(W\).}

To connect this distinction to method selection, we fix a benchmark distribution \(P\) over task--claim pairs \((z,c)\) and contexts \((O_z,C_c,H_z,E_c)\), independently of the evaluated method \(m\). For a method producing \(A_m(z;B,\omega)\), \(B\) is its budget and \(\omega\) follows the frozen reporting-seed protocol (degenerate for deterministic procedures); the two targets induce different estimands:
\[
\begin{aligned}
Q_m(B)
&=
\mathbb{E}_{(z,c)\sim P,\omega}
\!\left[q\!\left(A_m(z;B,\omega);R_z,K_z\right)\right],\\
S_m(B)
&=
\Pr_{(z,c)\sim P,\omega}
\!\left[W\!\left(A_m(z;B,\omega);O_z,C_c,H_z,E_c\right)=1\right].
\end{aligned}
\]
The difference lies in the paired endpoint being aggregated: \(Q_m\) aggregates artifact quality under the fixed criterion, whereas \(S_m\) aggregates support under the same frozen distribution and claim contexts. Maximizing \(Q_m\), rather than \(S_m\), can therefore reward different method properties and induce a different method selection.

In summary, Equation~\eqref{eq:evaluation-targets} formalizes an evaluation--use mismatch in AI for PDEs, and the two evaluation targets can induce different method selections. More importantly, because benchmarks shape which methods the community selects, improves, and treats as progress, this mismatch can propagate from evaluation to the direction of methodological development.

\section{Extending Evaluation from Accuracy to Evidential Support}
\label{sec:benchmark-extensions}
\raggedbottom

We extend evaluation from accuracy to evidential support on PDEBench and PDEInvBench, respectively: the former evaluates predicted fields or trajectories from forward simulation, whereas the latter evaluates parameters or coefficient fields recovered in inverse inference. We choose these two benchmarks because they combine public protocols, broad task coverage, reproducible baselines, and released references, allowing a claim-conditioned evaluator to be added without changing the underlying task~\citep{takamoto2022pdebench,goel2026pdeinvbench}.

We first specify how the evidential-support target is operationalized in these benchmark extensions. For a fixed object \(O\) and assumptions \(H\), we express a claim \(C\) as a prespecified acceptance set \(\mathcal C\) in a claim-relevant coordinate. The evidence standard \(E\) specifies how the artifact \(A\) and an independent calibration partition produce a region \(\Gamma^{0.95}\), and which containment relation constitutes support:
\[
 A \xrightarrow[\text{independent calibration}]{E} \Gamma^{0.95},
 \qquad
 W=1
 \Longleftrightarrow
 \bigl[\Gamma^{0.95}\text{ is valid}\bigr]
 \land
 \bigl[\Gamma^{0.95}\subseteq\mathcal C\bigr].
\]
Under exchangeability and finite scores, split-conformal calibration gives \(\Gamma^{0.95}\) finite-sample marginal coverage~\citep{lei2018distributionfree}, without access to the paired-evaluation reference. Given a valid region, a claim is \textsc{supported} inside \(\mathcal C\), \textsc{refuted} inside a prespecified disjoint complement, and \textsc{unknown} otherwise; no-region and invalid remain separate, and \(W=1\) only for \textsc{supported}. Appendix~\ref{app:claim-design-provenance} explains the claims' design principles, scientific provenance, and scope; Appendix~\ref{app:claims-evidence} gives full contract and coverage details.

With this operational definition in place, we extend PDEBench and PDEInvBench in Sections~\ref{sec:extend_pde} and~\ref{sec:extend_pdeinv}, respectively. In both extensions, \(q\) and \(W\) use the same pre-frozen eligible support, claim contracts, instances, submitted artifacts, seed protocols, and computational budgets; only the evaluation target changes.

\subsection{Extending Forward-Simulation Evaluation}\label{sec:extend_pde}

\paragraph{Original accuracy evaluation.}
PDEBench evaluates autoregressive prediction of future states for dynamic systems and coefficient-to-solution mapping for steady Darcy flow~\citep{takamoto2022pdebench}. In each case, the submitted field or trajectory is compared with the released numerical realization. RMSE, normalized RMSE, maximum error, and the applicable conservation, boundary, and frequency-band diagnostics characterize agreement with a numerical reference or tested constraint. We retain every applicable raw diagnostic separately and set the larger-is-better paired score to \(q=-\mathrm{nRMSE}\).

\paragraph{Claim-conditioned extension.}
We extend every audited PDEBench configuration lineage with two PDE-appropriate claim contracts, each instantiating \((O_z,C_c,H_z,E_c)\) together with a claim coordinate \(G_c\). These contracts cover four recurring claim categories in published scientific applications using AI-generated physical fields: regional quantities or loads, threshold exceedance or affected extent, oriented transport or flux, and event occurrence or timing~\citep{kochkov2024neuralgcm,yousefi2024drag,lam2023graphcast,zhang2023nowcastnet,bi2023pangu}. Table~\ref{tab:pdebench-claim-map} summarizes family-level applicability and the executed claim pairs. In the main text, we use two-dimensional shallow water as a concrete example. For Table~\ref{tab:pdebench-claim-instantiations}, the independent unit \(z\) is a complete trajectory, \(A_z=\widehat h_z(\cdot,\cdot)\) is the submitted height trajectory, and \(O_z=h_z^{\mathrm{ref}}(\cdot,\cdot)\) is the corresponding simulator-resolved object. The two claims share the physical and scope assumptions \(H_z^0\): the same benchmark cell, spatial domain, initial and boundary conditions, grid, units, and prediction horizon. Each row of the table then specifies the scientific claim \(C_c\), its claim coordinate \(G_c\), and the evidence standard \(E_c\).

\par\medskip
\noindent\begin{minipage}{\textwidth}
  \refstepcounter{table}
  \noindent\textbf{Table~\thetable:} Two shallow-water claim contracts evaluated on the same submitted artifact \(A_z\) and simulator-resolved object \(O_z\), under shared assumptions \(H_z^0\).\label{tab:pdebench-claim-instantiations}
  \par\smallskip
  \centering
  \footnotesize
  \setlength{\tabcolsep}{3.5pt}
  \renewcommand{\arraystretch}{1.12}
  \begin{tabularx}{\textwidth}{@{}>{\raggedright\arraybackslash}p{5.15cm}>{\raggedright\arraybackslash}p{2.85cm}>{\raggedright\arraybackslash}X@{}}
    \toprule
    Scientific claim \(C_c\) & Coordinate \(G_c(v)\) & Evidence standard \(E_c\) \\
    \midrule
    \(C_{\mathrm{mean}}\): the regional mean lies in the focal interval \([\ell_{\mathrm m},u_{\mathrm m}]\); the registered complements are \((-\infty,\ell_{\mathrm m}-\delta_{\mathrm m}]\) and \([u_{\mathrm m}+\delta_{\mathrm m},\infty)\), with the intervening bands left indeterminate
    &
    \(G_{\mathrm{mean}}(v)=|\mathcal D|^{-1}\!\int_{\mathcal D}v(x,t_*)\,dx\), for frozen \(\mathcal D\) and \(t_*\)
    &
    The registered four-way role and stratum calibrate absolute residuals in \(G_{\mathrm{mean}}\) to form \(\Gamma_{z,\mathrm{mean}}^{0.95}\); support requires a valid region inside the focal interval
    \\
    \(C_{\mathrm{extent}}\): the fraction above the registered level lies in \([\ell_{\mathrm e},u_{\mathrm e}]\); disjoint low and high complements are separated from it by the frozen gap \(\delta_{\mathrm e}\)
    &
    \(G_{\mathrm{extent}}(v)=|\mathcal D|^{-1}\!\int_{\mathcal D}\mathbf 1\{v(x,t_*)>b\}\,dx\), for frozen \(\mathcal D,t_*\), and level \(b\)
    &
    The same wrapper in \(G_{\mathrm{extent}}\) forms \(\Gamma_{z,\mathrm{extent}}^{0.95}\); support requires a valid region inside the focal interval
    \\
    \bottomrule
  \end{tabularx}
\end{minipage}
\par\medskip

To compare established and recent approaches across the benchmark's heterogeneous cells, we use FNO and U-Net as common classic anchors, PINNs in compatible low-dimensional cells, and the strong recent dynamics models DPOT, VCNeF, and OmniArch-B wherever eligible~\citep{takamoto2022pdebench,hao2024dpot,hagnberger2024vcnef,chen2025omniarch}. Thus, the PDEBench extension changes the evaluation endpoint from solution-field or rollout accuracy to evidential support for solution-derived scientific claims.

\subsection{Extending Inverse-Inference Evaluation}\label{sec:extend_pdeinv}

\paragraph{Original accuracy evaluation.}
At primary resolution, PDEInvBench maps observed solution windows or fields to five scalar targets across four dynamic systems---reaction--diffusion \(k\) and \(D_u\), unforced and forced Navier--Stokes viscosity, and KdV \(\delta\)---or a binary Darcy coefficient field, under named ID and OOD regimes~\citep{goel2026pdeinvbench}. Per-instance relative \(L_2\) error compares the recovered latent with its generating label; NLS remains the benchmark's separate scaling endpoint. We set \(q=-\mathrm{relative}\text{-}L_2\), without treating training or test-tuning residuals as a universal test endpoint.

\paragraph{Claim-conditioned extension.}
We extend every PDEInvBench target--regime cell with two target-appropriate claim contracts, each instantiating \((O_z,C_c,H_z,E_c)\) together with a claim coordinate \(G_c\). For scalar targets, these contracts concern physically meaningful ranges and directional thresholds or regimes; for Darcy, they concern regional phase fractions and contrasts. These forms reflect recurring scientific uses of recovered quantities to characterize physical ranges or regimes and heterogeneous regional structure~\citep{wang2025ice,zapf2022brain,riel2021glacier,huang2022permeability}. Table~\ref{tab:pdeinvbench-claim-map} summarizes target-level applicability and the registered claim pairs. In the main text, we use Darcy as a concrete example. For Table~\ref{tab:pdeinvbench-claim-instantiations}, the independent unit \(z\) is one Darcy field instance, \(A_z=\widehat a_z(\cdot)\) is the submitted recovered coefficient field, and \(O_z=a_z^*(\cdot)\) is the corresponding simulator-defined generating object. The two claims share the physical and scope assumptions \(H_z^0\): the same PDE and boundary conditions, grid, units, observation and noise model, and official regime. Each row of the table then specifies the scientific claim \(C_c\), its claim coordinate \(G_c\), and the evidence standard \(E_c\).

\par\medskip
\noindent\begin{minipage}{\textwidth}
  \refstepcounter{table}
  \noindent\textbf{Table~\thetable:} Two Darcy claim contracts evaluated on the same submitted artifact \(A_z\) and simulator-defined object \(O_z\), under shared assumptions \(H_z^0\).\label{tab:pdeinvbench-claim-instantiations}
  \par\smallskip
  \centering
  \footnotesize
  \setlength{\tabcolsep}{3.5pt}
  \renewcommand{\arraystretch}{1.12}
  \begin{tabularx}{\textwidth}{@{}>{\raggedright\arraybackslash}p{4.15cm}>{\raggedright\arraybackslash}p{4.05cm}>{\raggedright\arraybackslash}X@{}}
    \toprule
    Scientific claim \(C_c\) & Coordinate \(G_c(v)\) & Evidence standard \(E_c\) \\
    \midrule
    \(C_{\mathrm{frac}}\): the high-phase fraction in \(\mathcal D_1\) lies in \([\ell_1,u_1]\); disjoint low and high complements are separated from it by the frozen gap \(\delta_1\)
    &
    \(G_j(v)=|\mathcal D_j|^{-1}\!\int_{\mathcal D_j}\mathbf 1\{v(x)\geq\vartheta\}\,dx\), with \(j=1\), frozen high-phase convention, tolerance, \(\vartheta\), and mask \(\mathcal D_1\)
    &
    The registered four-way role and stratum calibrate absolute residuals in \(G_1\) to form \(\Gamma_{z,\mathrm{frac}}^{0.95}\); support requires a valid region inside the focal interval
    \\
    \(C_{\mathrm{contrast}}\): \(\mathcal D_1\) has the larger high-phase fraction, with focal member \([\delta_{\Delta},1]\), low complement \([-1,-\delta_{\Delta}]\), and indifference gap \((-\delta_{\Delta},\delta_{\Delta})\)
    &
    \(G_{\Delta}(v)=G_1(v)-G_2(v)\), for frozen ordered masks \((\mathcal D_1,\mathcal D_2)\) and high-phase map
    &
    The same wrapper in \(G_{\Delta}\) forms \(\Gamma_{z,\mathrm{contrast}}^{0.95}\); support requires a valid region inside the focal member
    \\
    \bottomrule
  \end{tabularx}
\end{minipage}
\par\medskip

To span amortized neural inversion, classical per-instance optimization, and recent target-specific or probabilistic operators, the scalar panel uses FNO, ResNet, and L-BFGS-B with ten starts alongside SC-FNO, CONFIDE, and FNOPE, while the Darcy panel uses FNO, ResNet, and DeepONet with iFNO, FNOPE, and DGenNO wherever eligible~\citep{goel2026pdeinvbench,behroozi2025scfno,linial2024confide,moss2025fnope,long2025ifno,zang2025dgenno}. Thus, the PDEInvBench extension changes the evaluation endpoint from scalar-parameter or coefficient-field recovery accuracy to evidential support for scientific claims about recovered parameters or coefficient-field functionals.
\flushbottom

\section{When and Why Accuracy and Evidential Support Rank Methods Differently}
\label{sec:different-rankings}

In this section, we answer three questions in three corresponding subsections: First, does the evidence evaluation behave as intended? Second, on an identical common support, do the original accuracy estimand \(Q_m\) and focal-support estimand \(S_m\) select the same method set? Third, can any difference be explained and predicted from measurable properties of the claim, artifact, and evidence region?

\subsection{Coverage, Informative Verdicts, and Artifact Sensitivity}
\label{sec:results-validity}

\textbf{Coverage is near nominal empirically.}
We first audit interval coverage and erroneous verdicts on the frozen ranked supports. Coverage is 94.6--96.8\% on PDEBench and 93.8--97.1\% on PDEInvBench; the all-assigned wrong-direction rates are 0.26\% and 0.54\%, and the corresponding invalid rates are 0.18\% and 0.49\%, respectively. Thus, the evidence regions show near-nominal empirical coverage and low directional error and invalidity on the supports analyzed below.

\textbf{The endpoint resolves clear cases.}
We next apply each claim contract to reference-consistent oracle artifacts in three registered regimes: a boundary-separated focal-true case, its boundary-separated complement, and a boundary-near case. All 70 PDEBench and 34 PDEInvBench contracts pass: within every contract, all 96 focal-true cases are supported and all 96 complement-true cases are refuted, with simultaneous worst-case lower bounds of 92.1\% and 92.8\%, respectively. By contrast, the boundary-near cases remain \textsc{unknown} at rates of 85.4--100\% and 83.3--100\%. Thus, the endpoint resolves both sides of a claim when an ideal artifact contains sufficient evidence while retaining uncertainty when the claim is not separated from its boundary; it is therefore non-vacuous.

\textbf{Verdicts respond to the submitted artifact.}
Finally, we replace the oracle artifact with null, random, shuffled, and task-swapped controls while leaving the claim and verifier fixed, and audit execution traces for prohibited reference exposure or solver calls. Oracle--control separation is 0.929--0.980 on PDEBench and 0.915--0.974 on PDEInvBench, with every simultaneous lower bound above 0.75; all 23 nominal tracks also pass the provenance and no-solver checks with zero prohibited exposures or calls. Aggregate denominators, contrasts, and diagnostic outcomes are reported in Appendix~\ref{app:ranking-mechanism-analysis}. Because resolution falls when task-relevant information in the artifact is destroyed, with no prohibited reference exposure or evaluator-side solver call, the added endpoint is artifact-dependent.

\subsection{Changing the Evaluation Target Changes Method Selection}
\label{sec:results-selection}

\begin{figure*}[t]
  \centering
  \includegraphics[width=\textwidth]{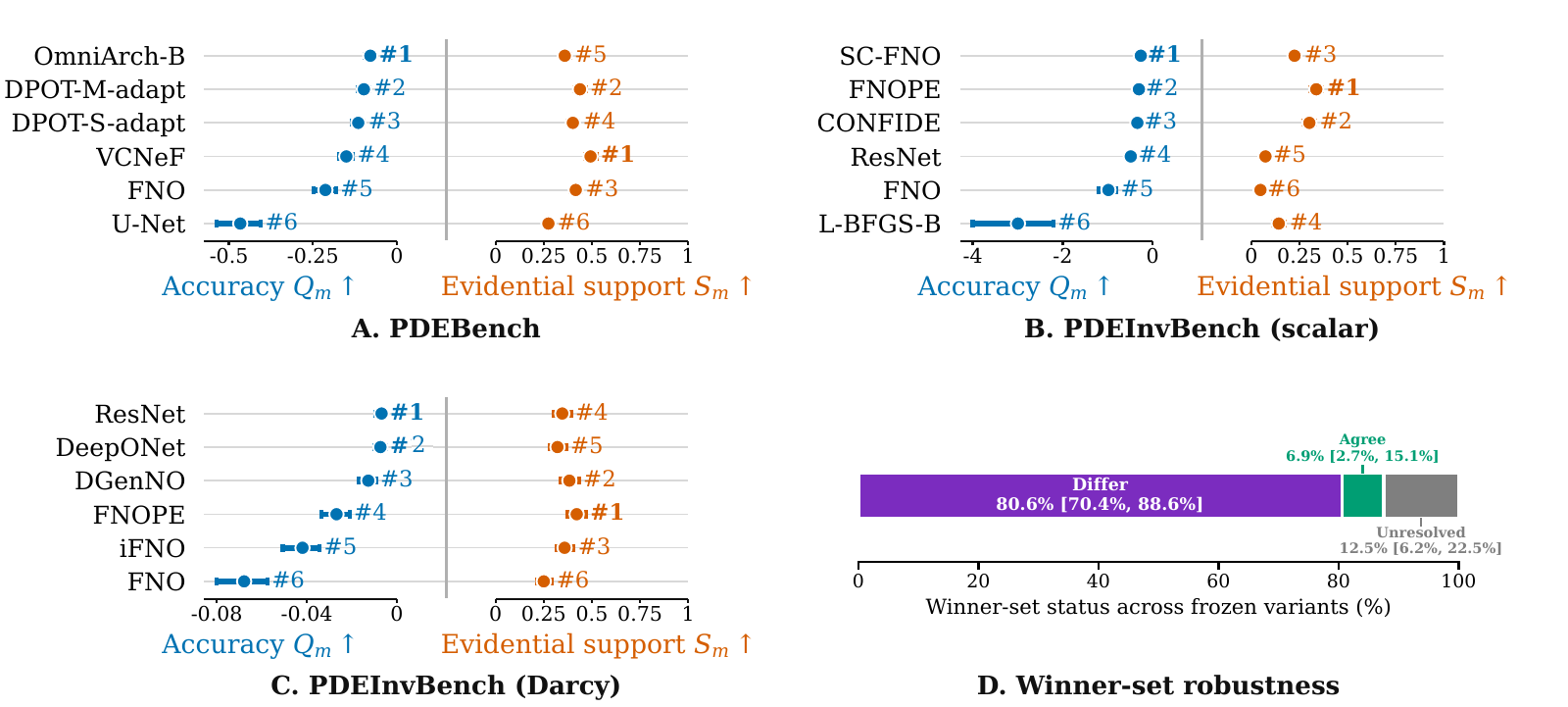}
  \caption{Paired \(Q/S\) leaderboards. Panels A--C show point-estimate ranks and simultaneous 95\% intervals; exact values and winner sets are in Appendix~\ref{app:sec5-selection}. Panel D summarizes winner-set robustness; Appendix~\ref{app:sec5-robustness} gives the full specification.}
  \label{fig:paired-selection-results}
\end{figure*}

Figure~\ref{fig:paired-selection-results} compares method selection under the two evaluation targets. Panels A--C show the aggregate \(Q\) and \(S\) leaderboards for PDEBench dynamics, inverse scalar inference, and inverse Darcy on their registered common supports. Panel D tests robustness by recomputing the nine primary records with disjoint \(Q\)- and \(S\)-winner sets under 72 frozen one-factor variants and recording whether the winner sets still differ, agree, or remain unresolved.

Across panels A--C, the two evaluation targets select different methods in all three headline comparisons. On PDEBench dynamics, \(Q\) selects OmniArch-B, whereas \(S\) selects VCNeF. On inverse scalar inference, \(Q\) selects SC-FNO whereas \(S\) selects FNOPE; on inverse Darcy, ResNet and FNOPE lead the \(Q\) and \(S\) point estimates, while DeepONet also belongs to the simultaneous \(Q\)-winner set. Across the 16 claim- and regime-specific primary records, the winner sets are disjoint in 9, match in 4, and remain unresolved or overlapping in 3. The separation is not merely a consequence of large accuracy differences: in the validation-nominated shallow-water comparison, DPOT-M and VCNeF are equivalent in global accuracy, \(\Delta Q=-0.00043\;[-0.00137,0.00054]\) within the frozen \(\pm0.003\) band, yet VCNeF has \(0.176\;[0.121,0.231]\) higher focal support.

Panel D shows that the winner sets remain disjoint in 58 of the 72 frozen variants (80.6\%; 95\% interval 70.4--88.6\%), agree in 5 (6.9\%; 2.7--15.1\%), and remain unresolved in 9 (12.5\%; 6.2--22.5\%). Hence, on identical common supports, changing the evaluation target changes the selected method set across most tested variants, although not universally.

\subsection{Boundary Proximity, Error Direction, and Evidence Width Predict Divergence}
\label{sec:results-mechanism}

For a one-sided claim in frozen stratum \(g=g(z,c)\), let \(\sigma_{zc}\in\{-1,1\}\) orient the accepted side, \(\tau_{zc}\) denote its boundary, and \(\widehat e_{mgcs}\) the calibrated region half-width. The support margin decomposes as
\begin{equation}
M_{mzsc}
=
\underbrace{\sigma_{zc}\!\left(G_c(R_z)-\tau_{zc}\right)}_{\text{reference-to-boundary clearance}}
+
\underbrace{\sigma_{zc}\!\left(G_c(A_{mzs})-G_c(R_z)\right)}_{\text{claim-directed error}}
-
\underbrace{\widehat e_{mgcs}}_{\text{evidence half-width}}.
\label{eq:boundary-direction-width}
\end{equation}
With \(T_{mzsc}\) denoting evidence-region status, support holds exactly under a valid region and nonnegative margin:
\begin{equation}
W_{mzsc}=1
\Longleftrightarrow
T_{mzsc}=\textsc{valid}\land M_{mzsc}\geq0.
\end{equation}
Global accuracy constrains error magnitude but not location, claim-relative direction, or evidence-region width. Equation~\eqref{eq:boundary-direction-width} therefore predicts greater agreement under large clearance, favorable claim-directed error, and tight evidence, and greater divergence under adverse joint geometry.

\begin{figure*}[t]
  \centering
  \includegraphics[width=\textwidth]{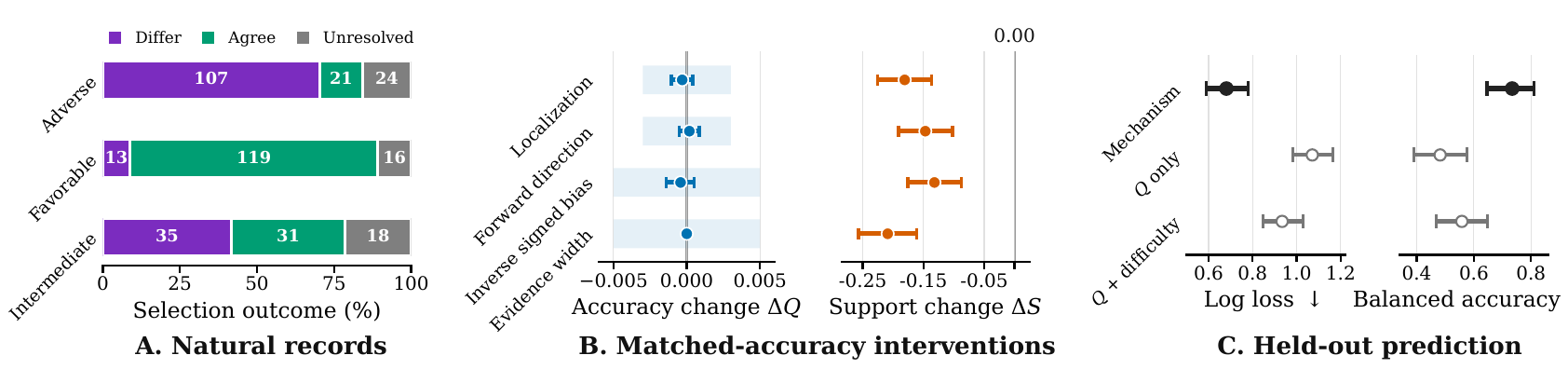}
  \caption{Evidence for the boundary--direction--width mechanism. Panel A shows selection outcomes across natural-record strata, panel B shows matched-accuracy interventions, and panel C compares held-out prediction with \(Q\)-based baselines. Points and lines in panels B--C denote estimates and 95\% intervals; shaded bands in panel B denote the frozen \(Q\)-equivalence bands.}
  \label{fig:mechanism-results}
\end{figure*}

As illustrated in Figure~\ref{fig:mechanism-results}, we test this prediction in three complementary ways. Panel A groups unmodified recomputations by validation-frozen joint geometry and asks whether selection outcomes shift between adverse and favorable strata. Panel B changes error localization or direction, inverse signed bias, or evidence width one factor at a time while keeping global accuracy within frozen equivalence bands, testing whether these factors can change support without a corresponding change in \(Q\). Panel C asks whether the same pre-verdict variables predict whether the two selections agree, differ, or remain unresolved on whole-family/target holdouts beyond \(Q\)-based baselines. Appendices~\ref{app:sec5-mechanism} and~\ref{app:sec5-prediction} give the full specifications and diagnostics.

The three panels support the predicted geometry at complementary levels. Descriptively, panel A shows that selections differ in 107/152 adverse-stratum recomputations but agree in 119/148 favorable-stratum recomputations, while the intermediate stratum has no dominant outcome. In panel B, every \(\Delta Q\) interval remains within its frozen equivalence band, yet adverse error localization, error direction, inverse signed bias, and wider evidence reduce focal support by \(0.181\;[0.137,0.226]\), \(0.147\;[0.102,0.191]\), \(0.132\;[0.088,0.176]\), and \(0.209\;[0.161,0.257]\), respectively. In panel C, across 52 whole-family/target holdouts, the mechanism model attains log loss \(0.681\;[0.588,0.780]\) and balanced accuracy \(0.734\;[0.646,0.811]\), outperforming \(Q\)-based baselines. Together, these results support the boundary--direction--width mechanism: divergence is more likely near the claim boundary, under adverse claim-directed error, or with wide evidence, whereas favorable geometry makes agreement more likely.

\section{Discussion}
\label{sec:discussion}

We first distinguish our contribution from benchmarks that pose open-ended scientific or research problems to LLM-based agents. For example, BLADE evaluates agents against expert analyses, DiscoveryBench scores recovered hypotheses and data-analysis workflows, and MLRC-Bench evaluates proposed research methods through competition outcomes~\citep{gu2024blade,majumder2025discoverybench,zhang2025mlrcbench}. Although these benchmarks test whether an LLM can produce a scientific conclusion or research outcome, they do not assess whether the evidence it generates is sufficient to support the corresponding conclusion. This distinction is becoming increasingly important as LLM-based agents can now generate hypotheses, analyses, and even complete research artifacts, while drawing reliable scientific conclusions from such outputs still requires rigorous verification~\citep{lu2026automation,cornelio2026verification}. These benchmarks therefore cannot answer our question: when does an AI-for-PDE output provide sufficient evidence for a specified scientific claim?

We directly answer this question in AI for PDEs by formalizing claim-conditioned evidential support, evaluating it alongside accuracy in two benchmark extensions, and explaining when and why the two endpoints select different methods. These findings matter because, when AI outputs are used as scientific evidence, accuracy-only leaderboards can reward methods whose outputs provide weaker support for the very claims those outputs are intended to support. Evidential support should therefore be evaluated as a distinct endpoint in its own right rather than inferred directly from accuracy.

Three limitations bound this conclusion. First, \(q(A;R,K)\) and \(W(A;O,C,H,E)\) are operational, not uniquely canonical; specifying the object, claim, assumptions, and evidence standard requires domain judgment and does not measure universal scientific truth. Second, evidence from two PDE benchmarks and finitely many claims and methods does not establish broader generalization across AI for science or real practice. Third, finite-score conformal guarantees are marginal under exchangeability, not individual-case or distribution-shift guarantees.

\section*{References}

\begingroup
\small
\renewcommand{\section}[2]{}

\endgroup


\clearpage
\appendix

\section{Experimental Design for the Benchmark Extensions}

\label{app:benchmark-extension-details}

This appendix records the design choices needed to interpret the benchmark extensions: the evaluated scope and independent units (Appendix A.1), the claim-design principles and scientific provenance (Appendix A.2), the frozen claim contracts and common evidence wrapper (Appendix A.3), the method-specific common supports and resource contracts (Appendix A.4), and the controls, endpoints, and statistical analysis (Appendix A.5). The experiment was completed on these frozen contracts.

\subsection{Benchmark Scope, Independent Units, and Data Roles}

\label{app:scope-splits}

\textbf{Evaluated scope.} The forward extension used the 35 audited, nonduplicate PDEBench DaRUS V8 configuration lineages in the sealed formal registry: 30 dynamic lineages spanning ten PDE families and five steady-Darcy lineages \citep{takamoto2022pdebench,takamoto2022pdebenchdata}. Each lineage was assigned two structurally applicable claim contracts, giving 70 PDEBench contracts. The inverse extension registered every primary-resolution PDEInvBench target and official primary regime: five scalar targets---reaction--diffusion \(k\) and \(D_u\), viscosity in unforced and forced Navier--Stokes, and KdV \(\delta\)---in ID, OOD non-extreme, and OOD extreme, plus the Darcy coefficient field in ID and OOD extreme \citep{goel2026pdeinvbench}. Two claims were assigned to each target--regime cell, giving \(5\times3\times2+1\times2\times2=34\) registered inverse contracts. Official availability leaves the six reaction--diffusion target--regime cells (12 contracts) structural \texttt{N/A\_PREOUTCOME}; complete-primary analysis therefore contains the other nine scalar cells and the two Darcy cells. Other released resolutions, except the registered forced-Navier--Stokes source binding described below, were excluded from nominal rankings and reserved for named sensitivity analyses; inherited-checkpoint tracks likewise remained secondary.

\begin{table}[!htbp]
\centering
\caption{Benchmark scope and independent units.}\label{tab:formal-benchmark-scope}
\small
\setlength{\tabcolsep}{2pt}
\begin{tabular}{@{}p{0.20\textwidth}p{0.21\textwidth}p{0.32\textwidth}p{0.19\textwidth}@{}}
\toprule
\begin{minipage}[b]{\linewidth}\raggedright
Component
\end{minipage} & \begin{minipage}[b]{\linewidth}\raggedright
Submitted artifact
\end{minipage} & \begin{minipage}[b]{\linewidth}\raggedright
Independent unit
\end{minipage} & \begin{minipage}[b]{\linewidth}\raggedleft
Registered / complete-primary contracts
\end{minipage} \\
\midrule
PDEBench dynamics & Predicted future field or trajectory & Complete trajectory & 30 lineages \(\times\) 2 claims = 60 \\
PDEBench Darcy & Predicted solution field & Coefficient--solution pair & 5 lineages \(\times\) 2 claims = 10 \\
PDEInvBench scalar & Recovered scalar parameter & Complete trajectory group, including all windows sharing an initial condition & 15 registered cells \(\times\) 2 claims = 30; 9 cells \(\times\) 2 claims = 18 complete-primary \\
PDEInvBench Darcy & Recovered coefficient field & Complete Darcy field instance & 2 regimes \(\times\) 2 claims = 4 \\
\bottomrule
\end{tabular}
\end{table}

\textbf{Independent assignment and reference separation.} All windows derived from one trajectory, and all trajectories linked by a reused initial condition, retained one group identifier and one role. Darcy coefficient--solution pairs and coefficient fields were assigned as whole units. The frozen split generator assigned each independent group to exactly one of four mutually exclusive roles: model fitting, validation, final calibration, or paired evaluation. Validation selected ordinary hyperparameters and the few design-pilot choices declared before the run; final calibration alone determined evidence-region radii; and paired evaluation alone supplied the primary accuracy/support comparison. Each complete-primary cell contributed 20 independent paired-evaluation groups, except forced Navier--Stokes OOD non-extreme, whose 32 official groups supplied 19 final-calibration and 13 paired-evaluation groups. Its registered \(64\!\times\!64\) fields were stride-32 coordinate subsets of the official \(2048\!\times\!2048\) fields, without interpolation or filtering. The separately constructed oracle and control bank was disjoint from these groups and never entered accuracy, support, coverage, or ranking denominators.

The common-wrapper worker had access to the submitted artifact, the frozen claim contract, and the sealed final-calibration quantity, but not to the paired-evaluation reference. It sealed the region and verdict before a separate audit joined the reference to compute benchmark accuracy, empirical coverage, or direction errors. This separation prevented the scientific claim, its boundary, or its verdict from being adapted to the evaluated answer.

\subsection{Claim-Design Principles and Scientific Provenance}
\label{app:claim-design-provenance}

This subsection explains why the 104 claim contracts are appropriate operational probes for the paper's conditional question, rather than arbitrary post hoc summaries of the benchmark outputs. It establishes four points in order. First, every frozen contract is auditable against common principles connecting a benchmark object to a scientifically interpretable and mechanically decidable claim. Second, the claim forms have direct provenance in published uses of AI-generated physical fields and recovered PDE parameters or fields, while their numerical bindings remain benchmark-specific operational choices. Third, every verdict-relevant design choice---including applicability, coordinate, boundary, score, stratum, shortage rule, and checker---was frozen before final calibration and paired evaluation; final calibration only supplied the evidence-region radius under those frozen rules, and the paired-evaluation reference was unavailable to the verdict worker. Fourth, the resulting suite covers every studied lineage or target--regime cell under a uniform assignment rule, without being presented as a complete or statistically representative taxonomy of scientific claims.

\subsubsection{Design Objective and Admissibility Principles}
\label{app:claim-design-principles}

\paragraph{Evidence collection and criteria.}
The evidence basis consists of two frozen records: the benchmark and PDE definitions, which identify the object, variables, units, and admissible structure; and the formal claim registry, which records the coordinate, acceptance sets, evidence construction, and applicability decision. The registry supplies these fields for all 104 contracts; Tables~\ref{tab:pdebench-claim-map} and~\ref{tab:pdeinvbench-claim-map} summarize family- and target-level structural applicability and the registered pairs, and Appendix~\ref{app:claims-evidence} specifies the common region construction and checker. We use these records to check five common constraints. \emph{Task fidelity} requires \(O_z\) and \(H_z\) to inherit the benchmark's physical object, PDE, units, observation scope, and independent unit. \emph{Scientific interpretability} requires \(G_c\) to extract a quantity for which a scoped statement about the object is meaningful under \(H_z\). \emph{Operational decidability} requires \(C_c\) and \(E_c\) to specify focal and complementary sets, any indifference gap, a calibrated region, and a deterministic checker that can return supported, refuted, or unknown. \emph{Structural applicability} requires the coordinate, masks, times, directions, and sets to be definable from the PDE and benchmark structure before paired outcomes are inspected. \emph{Method-independent evaluability} requires every eligible method to be judged from the same submitted artifact and wrapper, without a solver, inverse routine, or paired-evaluation reference inside the verdict worker.

These constraints make the evaluation chain explicit:
\[
\begin{aligned}
A
&\longrightarrow G_c(A)
\xrightarrow[\text{sealed final-calibration quantity}]{E_c}
\Gamma^{0.95}_{c}(A),\\
\left(
\Gamma^{0.95}_{c}(A),
\{\mathcal C_{c0},\mathcal C_{c1},\ldots\}
\right)
&\longrightarrow W_c(A)
\longrightarrow S_m
\longrightarrow \text{method selection}.
\end{aligned}
\]
Full-field accuracy and claim-conditioned support enter this chain at different places: \(q\) evaluates the artifact against a reference object or constraint, whereas \(W_c\) asks whether the calibrated region for a specified claim coordinate lies inside the claim's accepted set under the stated assumptions and evidence standard. Consequently, replacing field error by an arbitrary second field statistic would not satisfy the contract. \textbf{Conclusion:} each included contract is an atomic, auditable use of a benchmark artifact as evidence for a specified claim, rather than a relabeling of numerical accuracy.

\subsubsection{Scientific Provenance and Operational Binding}
\label{app:claim-scientific-provenance}

\paragraph{Evidence collection.}
For scientific provenance, we retained a published primary study only when the chain from an AI-generated physical field or recovered PDE quantity, through a claim-relevant summary, to a scientific interpretation was explicit. The purpose of this screening was to establish the provenance of the \emph{claim forms}; benchmark papers and PDE definitions were examined separately to establish which variables and structures can instantiate those forms. Sources that reported only full-field or parameter error, or that merely proposed a possible scientific use, were not used for this purpose.

\paragraph{Forward evidence.}
Published studies derive scientifically interpretable summaries from AI-generated physical fields rather than relying only on full-field error. NeuralGCM reports global climate quantities and annual regional tropical-cyclone counts, while Yousefi et~al. evaluate local and area-aggregate skin-friction loads~\citep{kochkov2024neuralgcm,yousefi2024drag}. GraphCast derives vertically integrated water-vapour transport, classifies temperature exceedances relative to fixed climatological thresholds, and extracts cyclone existence and trajectories; NowcastNet evaluates spatial rain areas defined by fixed rain-rate thresholds~\citep{lam2023graphcast,zhang2023nowcastnet}. Starting from known cyclone positions, Pangu-Weather applies a threshold-based tracker to forecast fields and evaluates landfall conclusions and landing time~\citep{bi2023pangu}. These uses motivate four forward templates: regional level or load, threshold exceedance or affected extent, oriented transport or flux, and event occurrence or time.

\paragraph{Inverse evidence.}
Scalar inverse studies interpret recovered physical parameters through reported numerical ranges~\citep{zapf2022brain}. Inferred coefficients, exponents, or directional changes are also used to distinguish physical regimes or mechanistic end members~\citep{wang2025ice,riel2021glacier}, while spatially resolved inverse fields are used to characterize heterogeneous local and regional structure~\citep{riel2021glacier,huang2022permeability}. These uses motivate the interval, threshold or regime, and regional-structure templates. For the Darcy contracts, the high/low phase convention and the phase-fraction coordinate additionally follow the PDEInvBench target itself: the coefficient field takes the values 3 and 12, and the benchmark uses the fraction of sites at the maximum value as a split statistic~\citep{goel2026pdeinvbench}.

The evidence separates four provenance layers. Published applications establish that a coordinate \emph{form} has a documented scientific use. The benchmark and PDE definitions establish the object, units, available variables, phase convention, and structural applicability. Fixed geometric or algebraic rules establish transformations, masks, times, and directions where the benchmark does not supply them directly. Finally, target-specific focal sets, complementary sets, and indifference gaps are operational bindings: where no domain- or benchmark-defined boundary exists, they were selected with fit or validation material under the frozen rules described below and are not presented as universal physical thresholds. \textbf{Conclusion:} the literature establishes scientific provenance for the forms, while the executed contracts are transparent, benchmark-specific operationalizations of those forms; the citations do not purport to validate every numerical boundary or mask.

\subsubsection{Prespecification and Separation from Evaluated Outcomes}
\label{app:claim-prespecification}

\paragraph{Evidence collection and audit trail.}
The frozen claim registry and configurations record the verdict-relevant design choices: applicability decisions, coordinate transforms, spatial masks, times, directions, physical or operational thresholds, focal and complementary sets, indifference gaps, calibration strata, scores, shortage rules, and checkers. These choices were fixed from PDE definitions, the scientific-use forms above, and declared fit or validation material before final calibration or paired evaluation. Final calibration alone determined the evidence-region radius under the frozen score and shortage rule; it could not change the claim population, coordinate, boundary, or checker. Boundaries were never centered on a paired-evaluation reference.

The information flow enforced the same separation at execution time. Each independent group had exactly one of four mutually exclusive roles: model fitting, validation, final calibration, or paired evaluation. The common-wrapper worker received the submitted artifact, the frozen claim contract, and the sealed final-calibration quantity, but not the paired-evaluation reference. It sealed the region and verdict before a separate audit joined the reference to compute benchmark accuracy, empirical coverage, direction errors, or rankings. The provenance and trace controls in Appendix~\ref{app:controls-gates} separately checked split leakage, reference exposure, evaluator recentering, prohibited reconstruction, and solver or inversion calls.

Fit and validation material could therefore inform the few declared design-pilot choices; the design is not claimed to be independent of those roles. The relevant protection is the absence of a feedback path from final-calibration or paired-evaluation outcomes to the claim population or its bindings. \textbf{Conclusion:} the claims were prespecified and frozen before the outcomes used in the reported accuracy--support comparison, so neither favorable paired references nor observed \(q\), \(W\), \(S_m\), or rankings could be used to redesign them.

\subsubsection{Coverage and Intended Interpretation}
\label{app:claim-coverage}

\paragraph{Evidence collection.}
Coverage was assessed against the sealed benchmark registry, not inferred from the number of application papers found. The registry contains 35 audited PDEBench configuration lineages and 17 PDEInvBench target--regime cells. The fixed assignment rule gives every lineage or cell two structurally applicable claims, yielding 70 forward and 34 inverse contracts. Tables~\ref{tab:pdebench-claim-map} and~\ref{tab:pdeinvbench-claim-map} summarize the family- and target-level applicability classes and registered pairs; the sealed registry and frozen configurations record the lineage- and target--regime-cell resolutions and their numerical bindings.

\begin{table}[!htbp]
\centering
\caption{Complete PDEBench family-level structural applicability map and executed primary pair. A = structurally applicable throughout the family, C = resolved by a lineage-level audit before calibration, and N/A = structurally inapplicable.}\label{tab:pdebench-claim-map}
\small
\setlength{\tabcolsep}{2pt}
\begin{tabular}{@{}p{0.15\textwidth}p{0.13\textwidth}p{0.14\textwidth}p{0.13\textwidth}p{0.13\textwidth}p{0.24\textwidth}@{}}
\toprule
\begin{minipage}[b]{\linewidth}\raggedright
PDEBench family
\end{minipage} & \begin{minipage}[b]{\linewidth}\centering
Regional quantity
\end{minipage} & \begin{minipage}[b]{\linewidth}\centering
Threshold/\allowbreak extent
\end{minipage} & \begin{minipage}[b]{\linewidth}\centering
Flux/transport
\end{minipage} & \begin{minipage}[b]{\linewidth}\centering
Event/time
\end{minipage} & \begin{minipage}[b]{\linewidth}\raggedright
Executed primary pair
\end{minipage} \\
\midrule
1D Advection & A & A & A & A & Regional + frozen threshold/event \\
1D Burgers & A & A & A & A & Regional + frozen threshold/event \\
1D Diffusion--Reaction & C & C & C & C & Regional + frozen threshold/event \\
1D Diffusion--Sorption & A & A & C & A & Regional + frozen threshold/event \\
2D Diffusion--Reaction & C & C & C & C & Regional + frozen threshold/event \\
1D Compressible Navier--Stokes & C & C & C & C & Regional + frozen threshold/event \\
2D Compressible Navier--Stokes & C & C & C & C & Regional + frozen threshold/event \\
3D Compressible Navier--Stokes & C & C & C & C & Regional + frozen threshold/event \\
2D Incompressible Navier--Stokes & C & C & C & C & Regional + frozen threshold/event \\
2D Shallow Water & A & A & N/A & A & Regional mean height + exceedance extent \\
2D Darcy & A & A & A & N/A & Regional solution functional + oriented flux \\
\bottomrule
\end{tabular}
\end{table}

\begin{table}[!htbp]
\centering
\caption{Complete PDEInvBench target-level structural applicability map and registered claim pair. A = structurally applicable, C = conditional on a target--regime audit completed before calibration, and N/A = structurally inapplicable.}\label{tab:pdeinvbench-claim-map}
\small
\setlength{\tabcolsep}{2pt}
\begin{tabular}{@{}p{0.15\textwidth}p{0.13\textwidth}p{0.14\textwidth}p{0.13\textwidth}p{0.13\textwidth}p{0.24\textwidth}@{}}
\toprule
\begin{minipage}[b]{\linewidth}\raggedright
PDEInvBench target
\end{minipage} & \begin{minipage}[b]{\linewidth}\centering
Interval
\end{minipage} & \begin{minipage}[b]{\linewidth}\centering
Threshold/\allowbreak regime
\end{minipage} & \begin{minipage}[b]{\linewidth}\centering
Phase fraction
\end{minipage} & \begin{minipage}[b]{\linewidth}\centering
Regional contrast
\end{minipage} & \begin{minipage}[b]{\linewidth}\raggedright
Registered claim pair
\end{minipage} \\
\midrule
Reaction--diffusion \(k\) & A & A & N/A & N/A & Interval + directional threshold \\
Reaction--diffusion \(D_u\) & A & C & N/A & N/A & Interval + audited directional threshold \\
Unforced Navier--Stokes viscosity & A & A & N/A & N/A & Interval + viscosity-regime threshold \\
Forced Navier--Stokes viscosity & A & A & N/A & N/A & Interval + viscosity-regime threshold \\
KdV \(\delta\) & A & C & N/A & N/A & Interval + audited directional threshold \\
Darcy coefficient field & N/A & N/A & A & A & Regional high-phase fraction + ordered contrast \\
\bottomrule
\end{tabular}
\end{table}

The reaction--diffusion pairs remain registered, but their six target--regime cells are structural \texttt{N/A\_PREOUTCOME} for primary analysis; the other 11 inverse cells enter complete primary.

The tables establish complete coverage only relative to the studied registry and the two-claim assignment rule. They do not make the suite exhaustive over possible scientific questions, canonical for the represented PDEs, or statistically representative of AI-for-PDE practice. Each \(W_c\) is also an atomic verdict for one object, claim, assumption set, and evidence standard; composing multiple claims or heterogeneous evidence sources into a paper-level scientific conclusion lies outside the present evaluation. A \textsc{supported} verdict therefore means that the artifact meets the prespecified evidence standard for that claim under \(H_z\), not that the claim has been established as universally true.

This bounded coverage is sufficient for the paper's empirical question. The paired design tests whether accuracy- and support-based selections can diverge on the studied benchmarks and whether that divergence follows the proposed boundary--direction--width mechanism; it does not estimate how often such divergence occurs across all scientific uses of AI for PDEs. In conclusion, the suite is complete for the declared benchmark registry under its assignment rule and scientifically grounded at the level of claim form, but it is intentionally an operational test suite rather than a universal claim taxonomy.

\subsection{Frozen Claim Contracts and the Common Evidence Wrapper}
\label{app:claims-evidence}

\textbf{Contract representation.} Each contract was frozen as \((O_z,C_c,H_z,E_c)\) together with a claim coordinate \(G_c\). Here \(O_z\) is the physical object linked to independent unit \(z\); \(H_z\) records the PDE, units, observation scope, and exchangeability assumptions; \(G_c\) extracts the scientific quantity addressed by the claim; \(C_c\) gives a focal acceptance set, pairwise-disjoint complements, and any indifference gap; and \(E_c\) specifies applicability, calibration stratum, score, region construction, and checker. Several claims could use one artifact, but they retained the same independent-unit identifier and therefore did not create additional calibration or bootstrap observations.

\textbf{Common 95\% wrapper.} For method \(m\), seed \(s\), claim \(c\), and independent final-calibration group \(\mathcal Z_i\) in frozen stratum \(g\), the primary score was the group maximum of the absolute claim-coordinate residual,

\[
r_{migcs}
=
\max_{z\in\mathcal Z_i}
\left|G_c(A_{mzs})-G_c(R_z)\right|,
\]

with an extended value of \(+\infty\) when a required calibration artifact was invalid. For \(n_g\) independent calibration groups and \(\alpha=0.05\), we set

\[
k_g=\left\lceil(n_g+1)(1-\alpha)\right\rceil,
\qquad
\widehat e_{mgcs}=r_{m(k_g)gcs},
\]

and, when \(k_g\le n_g\) and the selected radius was finite, returned

\[
\Gamma^{0.95}_{mzsc}
=
\left[G_c(A_{mzs})-\widehat e_{mgcs},
      G_c(A_{mzs})+\widehat e_{mgcs}\right].
\]

Every complete-primary claim stratum contained exactly 19 independent final-calibration groups, so \(k_g=\lceil 0.95\times20\rceil=19\).

A primary stratum without the 19 independent groups required by this finite 95\% order statistic returned \textbf{no region} under the frozen shortage rule; no sibling-regime or post-outcome pooling was permitted. Under exchangeability and finite scores, the construction provides marginal coverage for a future independent group in the named stratum. Operational no-region and invalid cases were instead counted as uncovered in the empirical all-assigned diagnostic; no conditional, per-instance, or automatic OOD coverage is claimed \citep{lei2018distributionfree}.

\textbf{Verdicts.} For a valid region, the checker returned \textbf{supported} if \(\Gamma^{0.95}_{mzsc}\subseteq\mathcal C_{c0}\), \textbf{refuted} if it lay wholly inside one complementary member frozen before final calibration, and \textbf{unknown} otherwise. An absent finite radius yielded \textbf{no region}, while malformed, nonfinite, or interface-incompatible artifacts were \textbf{invalid} rather than epistemically unknown. The focal-support endpoint was

\[
W_{mzsc}
=
\mathbf 1\!\left\{
\Gamma^{0.95}_{mzsc}\mbox{ is valid and }
\Gamma^{0.95}_{mzsc}\subseteq\mathcal C_{c0}
\right\}.
\]

Thus refuted, unknown, no-region, and invalid assignments contributed zero without being conflated in diagnostic reporting. Native posteriors or intervals were calibrated through nested region families frozen before final calibration and evaluated by the same containment checker, but remained a separate evidence track. The 90\% standard, scale-normalized scores, and joint product regions were likewise sensitivity analyses and never replaced the primary 95\% absolute-residual wrapper.

\subsection{Methods, Common Supports, and Resource Contracts}

\label{app:methods-budgets}

\textbf{Frozen common supports.} Eligibility was decided before evaluation from method identity, public implementation and license, input/output and observation interfaces, spatial dimension, required solver or gradient access, checkpoint provenance, adapter specification, schema smoke tests, and a frozen resource cap---never from performance. A primary comparison used the exact intersection of cells, claims, data roles, artifact interfaces, and budgets shared by every method in that record. N/A cells were not imputed, and an eligible timeout, numerical failure, or invalid artifact remained visible rather than shrinking the support.

\begin{table}[!htbp]
\centering
\small
\setlength{\tabcolsep}{2pt}
\begin{tabular}{@{}p{0.21\textwidth}p{0.26\textwidth}p{0.45\textwidth}@{}}
\toprule
\begin{minipage}[b]{\linewidth}\raggedright
Support ID
\end{minipage} & \begin{minipage}[b]{\linewidth}\raggedright
Frozen scope
\end{minipage} & \begin{minipage}[b]{\linewidth}\raggedright
Compared methods
\end{minipage} \\
\midrule
\texttt{PDB-DYN-C6-30L} & 30 dynamic lineages in ten PDE families & FNO, U-Net, DPOT-S-adapt, DPOT-M-adapt, VCNeF, OmniArch-B \\
\texttt{PDB-PINN-C3-13L} & 13 released-interface dynamic lineages & FNO, U-Net, PINN \\
\texttt{PDB-DARCY-C2-5L} & Five steady-Darcy lineages & FNO, U-Net \\
\texttt{INV-SCALAR-C6-15R} & Five registered scalar targets in 15 cells; complete primary: three targets in nine cells & FNO, ResNet, SC-FNO, CONFIDE, FNOPE, ten-start L-BFGS-B \\
\texttt{INV-DARCY-C6-2R} & Darcy ID and OOD-extreme & FNO, ResNet, DeepONet, iFNO, FNOPE, DGenNO \\
\bottomrule
\end{tabular}
\end{table}

The dynamic panel used fit-only retraining of DPOT-S/M with frozen dimension/channel/grid adapters, fit-only OmniArch-B without Aligner, and base VCNeF rather than VCNeF-R \citep{hao2024dpot,hagnberger2024vcnef,chen2025omniarch}. DPOT-S and DPOT-M were two registered sizes within one DPOT architecture slot; with VCNeF and OmniArch-B, they formed the three recent architecture families on this support. The PINN comparison was restricted to its 13 supported lineages instead of being inserted into the six-method dynamic panel. The recent inverse methods entered only through audited target-specific adapters \citep{behroozi2025scfno,linial2024confide,moss2025fnope,long2025ifno,zang2025dgenno}; both the original accuracy metric and the common wrapper acted on each method's frozen decoded physical point functional, while samples and native regions were used only in the separately labeled native-evidence track.

Every learned method--cell pair used three reporting seeds with identical data roles. The deterministic L-BFGS-B anchor used its frozen ten-start protocol and was not assigned artificial seeds. Nominal checkpoints were trained or adapted without final-calibration or paired-evaluation identifiers; released checkpoints with unexcluded or unknown upstream exposure remained in labeled secondary tracks. This preserves method identity while preventing inherited data exposure from benefiting a nominal support ranking.

Fairness meant holding fixed the task, paired instances, artifact used by the two evaluators, data roles, evidence wrapper, seed protocol, eligible support, and declared budget frontier. It did not mean forcing methods with different legitimate interfaces to consume identical resource types. Resources were therefore frozen and reported componentwise as

\[
B=(B_{\mathrm{train}},B_{\mathrm{infer}},B_{\mathrm{sim}},B_{\mathrm{cal}},B_{\mathrm{verify}}),
\]

covering fitting, artifact generation, simulator/gradient/test-time calls, calibration, and claim verification. This separates, for example, simulator-based posterior training from amortized inference and prevents inherited compute from being silently recorded as zero.

\subsection{Controls, Endpoints, Statistical Analysis, and Release}

\label{app:controls-gates}

\textbf{Controls and diagnostics.} The validity experiment used three oracle regimes for each of the 104 claim contracts: 96 boundary-separated focal-true groups, 96 boundary-separated complement-true groups, and 48 boundary-near groups. For each of the 12 availability-limited reaction--diffusion contracts, the control audit used a separate prespecified 19-group calibration population from the registered simulator; these groups never entered primary denominators. This tested whether the endpoint could resolve either side when ideal artifacts were informative while remaining uncertain near a claim boundary. Null, random, within-task shuffled, and task-swapped artifacts passed through the identical interface to test whether verdicts could be produced without instance-specific information. Artifact corruptions tested sensitivity to submitted content; provenance and trace audits checked for split leakage, reference exposure, evaluator recentering, prohibited reconstruction, or solver/inversion calls. Native-region dispersion perturbations were confined to the secondary native track.

Coverage, wrong-direction resolution, and invalid rate were reported for every ranked support. Oracle attainability, oracle--control specificity, and provenance/no-solver status were checked before interpreting those supports. Application-level simultaneous bounds were retained as conservative gate diagnostics; threshold misses withheld the corresponding application-level calibration interpretation but did not alter raw \(W\) or pooled comparisons. Benchmark-pooled all-assigned rates were descriptive summaries, not substitutes for those gate decisions. No diagnostic could delete a lineage, target, seed, or unfavorable record, or overwrite unrelated method verdicts. Appendix B reports the completed outcomes.

\textbf{Endpoints.} \label{app:endpoints-statistics}

The original paired endpoint was oriented so that larger is better:

\[
q^{\mathrm{fwd}}=-\mathrm{nRMSE}_{\mathrm{PDEBench}},
\qquad
q^{\mathrm{inv}}=-\mathrm{relative}\mbox{-}L_2.
\]

All other benchmark-native errors and diagnostics, including PDEInvBench NLS, were reported separately and were not combined into a heterogeneous composite.

The evidential endpoint \(S_m\) was the frozen macro-average of \(W\) on the same artifacts and common support. All assigned units remained in its denominator; unknown, no-region, invalid, and missing-artifact states could not improve support through abstention. If an assigned artifact lacked finite \(q\), its comparison record was unavailable for \(Q_m\) rather than reduced to an easier complete-case support. We additionally reported artifact completion, correct- and wrong-direction resolution, empirical all-assigned coverage, invalid rate, calibrated-region width, componentwise cost, and the signed set-to-boundary margin

\[
M(\Gamma,\mathcal C)
=
\inf_{x\in\Gamma}
\left[d(x,\mathcal C^{\mathsf c})-d(x,\mathcal C)\right],
\]

which is nonnegative exactly when a valid region is contained in the accepted set (up to the registered boundary convention). These diagnostics explain support but were not combined into a new claim-independent scalar leaderboard.

\textbf{Aggregation and uncertainty.} Atomic observations were first reduced within each independent trajectory, reused-initial-condition group, or Darcy field. Multiple claims, thresholds, time points, pixels, or overlapping windows derived from one artifact retained one group identifier and never increased the effective sample size. PDEBench estimates then averaged across applicable claims within a lineage and macro-averaged PDE families equally; PDEInvBench estimates averaged claims and regimes within a target and macro-averaged targets equally.

All endpoint comparisons were paired on the frozen common support. With 10,000 draws and seed 0, the hierarchical bootstrap resampled PDE families, lineages within family, and independent groups within lineage for PDEBench; it resampled inverse targets, regimes within target, and independent groups within regime for PDEInvBench. Reporting seeds 0, 1, and 2 were jointly resampled across methods without being treated as independent scientific cases; deterministic procedures remained fixed. Simultaneous max-deviation intervals defined endpoint-specific winner confidence sets, avoiding an arbitrary tie-breaker: disjoint sets established different selections, identical winner confidence sets established matching selections, and overlapping sets remained unresolved.

The robustness analysis changed one frozen factor at a time---claim boundary, evidence construction or level, aggregation, reporting seed or held-out family/target, method eligibility, or budget frontier---while holding the rest of the record fixed. Mechanism interventions altered error location, direction, or evidence width subject to frozen accuracy-equivalence and feasibility checks. Held-out prediction kept every method, artifact, claim, and seed from an entire PDE family or inverse target in the same outer fold; its features were limited to design records, validation-accuracy profiles, and final-calibration summaries available before paired-evaluation verdicts. This prevented the proposed mechanism from being evaluated on the same outcome information used to define it.

\textbf{Code availability.} \label{app:reproducibility}

The accompanying repository provides the implementation, environment specifications, analysis code, and frozen benchmark, split, claim, method, control, intervention, and analysis configurations, including lineage- and cell-specific numerical claim bindings. It provides commands that generate machine-readable tables and figures from completed production records; benchmark data, runs, checkpoints, and aggregate paper outcomes are not distributed with the source package.

\setcounter{table}{0}
\section{Additional Experimental Results}

\label{app:ranking-mechanism-analysis}

Appendix A specifies the benchmark scope and data roles, claim-design principles and scientific provenance, frozen claim contracts and common evidence wrapper, method supports and resource contracts, and admissibility controls and statistical analysis. This appendix reports the additional results needed to support the three conclusions in Section 5: the added endpoint is valid, changing the endpoint changes method selection, and the resulting divergence follows a predictable boundary--direction--width mechanism.

\subsection{Endpoint Validity}

\label{app:sec5-validity}

Across the 104 contracts and controls specified in Appendix~\ref{app:controls-gates}, the completed validity audits yielded the following results.

\begin{table}[!htbp]
\centering
\small
\setlength{\tabcolsep}{2pt}
\begin{tabular}{@{}p{0.14\textwidth}p{0.31\textwidth}p{0.31\textwidth}p{0.17\textwidth}@{}}
\toprule
\begin{minipage}[b]{\linewidth}\raggedright
Audit
\end{minipage} & \begin{minipage}[b]{\linewidth}\raggedright
PDEBench
\end{minipage} & \begin{minipage}[b]{\linewidth}\raggedright
PDEInvBench
\end{minipage} & \begin{minipage}[b]{\linewidth}\raggedright
Outcome
\end{minipage} \\
\midrule
Oracle attainability & Every contract has 96/96 supported focal-true and 96/96 refuted complement-true separated cases; simultaneous worst-case lower bound 92.1\%; boundary-near unknown range 41/48--48/48 & Every contract has 96/96 in both separated directions; lower bound 92.8\%; boundary-near unknown range 40/48--48/48 & All 104 contracts pass \\
Specificity & On 13,440 records per control, correct-resolution counts are 958, 722, 489, and 267; oracle--control separations are \(0.929\,[0.902,0.951]\), \(0.946\,[0.919,0.967]\), \(0.964\,[0.938,0.979]\), and \(0.980\,[0.955,0.990]\) & On 6,528 records per control, counts are 558, 397, 291, and 169; separations are \(0.915\,[0.879,0.941]\), \(0.939\,[0.908,0.961]\), \(0.955\,[0.927,0.975]\), and \(0.974\,[0.949,0.987]\) & Every simultaneous lower bound exceeds 0.75 \\
Calibration and direction & Coverage 94.6--96.8\%; all-assigned wrong-direction 63/24,360 (0.259\%); invalid 44/24,360 (0.181\%) & Coverage 93.8--97.1\%; all-assigned wrong-direction 38/6,976 (0.545\%); invalid 34/6,976 (0.487\%) & Near-nominal coverage; low aggregate error rates \\
Provenance & Zero prohibited exposures or evaluator-side solver calls; 11/11 nominal tracks pass & Zero prohibited exposures or calls; 12/12 nominal tracks pass & 23/23 nominal tracks pass \\
\bottomrule
\end{tabular}
\end{table}

The audit contains 24,360 unique PDEBench and 6,976 PDEInvBench submitted \(W\) rows (31,336 total). Wrong-direction resolutions (63 and 38) and invalid rows (44 and 34) remain in the all-assigned denominators. Seven native/corruption tracks are treated as secondary diagnostics, eight registered inherited-checkpoint assignments with unresolved upstream-corpus exposure are excluded from nominal rankings, and no trace-invalid track enters a ranking. Thus the endpoint resolves separated oracle cases, becomes uncertain near the registered boundary, loses resolution when task-relevant artifact information is destroyed, shows near-nominal empirical coverage and low aggregate direction-error and invalidity rates, and passes its provenance checks.

\subsection{Paired Leaderboards and Primary Selections}

\label{app:sec5-selection}

Using the paired endpoints and simultaneous winner-confidence-set procedure specified in Appendix~\ref{app:endpoints-statistics}, the three headline leaderboards are reported below. Rank columns follow point estimates; selections use the simultaneous winner confidence sets.

\begin{table}[!htbp]
\centering
\scriptsize
\setlength{\tabcolsep}{2pt}
\begin{tabular}{@{}p{0.12\textwidth}p{0.14\textwidth}p{0.25\textwidth}p{0.07\textwidth}p{0.25\textwidth}p{0.07\textwidth}@{}}
\toprule
\begin{minipage}[b]{\linewidth}\raggedright
Benchmark
\end{minipage} & \begin{minipage}[b]{\linewidth}\raggedright
Method
\end{minipage} & \begin{minipage}[b]{\linewidth}\raggedleft
\(Q\) (simultaneous 95\% interval)
\end{minipage} & \begin{minipage}[b]{\linewidth}\raggedleft
\(Q\) point rank
\end{minipage} & \begin{minipage}[b]{\linewidth}\raggedleft
\(S\) (simultaneous 95\% interval)
\end{minipage} & \begin{minipage}[b]{\linewidth}\raggedleft
\(S\) point rank
\end{minipage} \\
\midrule
PDEBench dynamics & OmniArch-B & \(-0.07902\,[-0.0946,-0.0657]\) & 1 & \(0.359\,[0.333,0.386]\) & 5 \\
& DPOT-M-adapt & \(-0.09816\,[-0.1159,-0.0827]\) & 2 & \(0.438\,[0.410,0.467]\) & 2 \\
& DPOT-S-adapt & \(-0.11505\,[-0.1346,-0.0982]\) & 3 & \(0.401\,[0.372,0.430]\) & 4 \\
& VCNeF & \(-0.15048\,[-0.1728,-0.1312]\) & 4 & \(0.493\,[0.463,0.524]\) & 1 \\
& FNO & \(-0.21266\,[-0.2489,-0.1814]\) & 5 & \(0.416\,[0.387,0.445]\) & 3 \\
& U-Net & \(-0.46611\,[-0.5357,-0.4052]\) & 6 & \(0.274\,[0.251,0.298]\) & 6 \\
Inverse scalar & SC-FNO & \(-0.26368\,[-0.318,-0.216]\) & 1 & \(0.224\,[0.198,0.251]\) & 3 \\
& FNOPE & \(-0.30562\,[-0.361,-0.257]\) & 2 & \(0.337\,[0.306,0.368]\) & 1 \\
& CONFIDE & \(-0.34040\,[-0.403,-0.286]\) & 3 & \(0.301\,[0.272,0.331]\) & 2 \\
& ResNet & \(-0.48638\,[-0.572,-0.414]\) & 4 & \(0.073\,[0.057,0.091]\) & 5 \\
& FNO & \(-0.98392\,[-1.206,-0.807]\) & 5 & \(0.047\,[0.034,0.061]\) & 6 \\
& Ten-start L-BFGS-B & \(-2.99580\,[-4.002,-2.207]\) & 6 & \(0.142\,[0.112,0.175]\) & 4 \\
Inverse Darcy & ResNet & \(-0.0068\,[-0.0094,-0.0049]\) & 1 & \(0.346\,[0.304,0.391]\) & 4 \\
& DeepONet & \(-0.0073\,[-0.0101,-0.0050]\) & 2 & \(0.321\,[0.281,0.365]\) & 5 \\
& DGenNO & \(-0.0127\,[-0.0168,-0.0093]\) & 3 & \(0.383\,[0.339,0.432]\) & 2 \\
& FNOPE & \(-0.0268\,[-0.0334,-0.0210]\) & 4 & \(0.421\,[0.373,0.469]\) & 1 \\
& iFNO & \(-0.0419\,[-0.0507,-0.0343]\) & 5 & \(0.358\,[0.315,0.405]\) & 3 \\
& FNO & \(-0.0678\,[-0.0799,-0.0575]\) & 6 & \(0.250\,[0.214,0.293]\) & 6 \\
\bottomrule
\end{tabular}
\end{table}

The corresponding simultaneous selections differ at all three headline levels: PDEBench dynamics selects \(\{\mbox{OmniArch-B}\}\) under \(Q\) and \(\{\mbox{VCNeF}\}\) under \(S\); inverse scalar selects \(\{\mbox{SC-FNO}\}\) and \(\{\mbox{FNOPE}\}\); and inverse Darcy selects \(\{\mbox{ResNet},\mbox{DeepONet}\}\) and \(\{\mbox{FNOPE}\}\). The 16 claim- and regime-specific records show where this aggregate conclusion holds and where the data do not resolve a difference.

\begin{table}[!htbp]
\centering
\scriptsize
\setlength{\tabcolsep}{2pt}
\begin{tabular}{@{}p{0.23\textwidth}p{0.24\textwidth}p{0.29\textwidth}p{0.16\textwidth}@{}}
\toprule
\begin{minipage}[b]{\linewidth}\raggedright
Record
\end{minipage} & \begin{minipage}[b]{\linewidth}\raggedright
\(Q\)-winner confidence set
\end{minipage} & \begin{minipage}[b]{\linewidth}\raggedright
\(S\)-winner confidence set
\end{minipage} & \begin{minipage}[b]{\linewidth}\raggedright
Comparison
\end{minipage} \\
\midrule
\texttt{PDB-DYN-C6-RGNL} & \{OmniArch-B\} & \{VCNeF\} & Disjoint \\
\texttt{PDB-DYN-C6-EVNT} & \{OmniArch-B\} & \{VCNeF\} & Disjoint \\
\texttt{PDB-PINN-C3-RGNL} & \{FNO\} & \{FNO\} & Match \\
\texttt{PDB-PINN-C3-EVNT} & \{FNO\} & \{FNO, PINN\} & Unresolved overlap \\
\texttt{PDB-DARCY-C2-REGN} & \{U-Net\} & \{U-Net\} & Match \\
\texttt{PDB-DARCY-C2-FLUX} & \{U-Net\} & \{U-Net\} & Match \\
\texttt{INV-SCALAR-ID-INT} & \{SC-FNO\} & \{FNOPE\} & Disjoint \\
\texttt{INV-SCALAR-ID-THR} & \{SC-FNO\} & \{CONFIDE\} & Disjoint \\
\texttt{INV-SCALAR-OODM-INT} & \{SC-FNO\} & \{FNOPE\} & Disjoint \\
\texttt{INV-SCALAR-OODM-THR} & \{SC-FNO\} & \{SC-FNO\} & Match \\
\texttt{INV-SCALAR-OODX-INT} & \{SC-FNO, FNOPE\} & \{SC-FNO, FNOPE, CONFIDE\} & Unresolved overlap \\
\texttt{INV-SCALAR-OODX-THR} & \{SC-FNO, FNOPE\} & \{CONFIDE\} & Disjoint \\
\texttt{INV-DARCY-ID-FRAC} & \{ResNet, DeepONet\} & \{FNOPE\} & Disjoint \\
\texttt{INV-DARCY-ID-CON} & \{ResNet, DeepONet\} & \{DGenNO\} & Disjoint \\
\texttt{INV-DARCY-OODX-FRAC} & \{ResNet, DeepONet, DGenNO\} & \{DGenNO, FNOPE\} & Unresolved overlap \\
\texttt{INV-DARCY-OODX-CON} & \{ResNet, DeepONet, DGenNO\} & \{FNOPE\} & Disjoint \\
\bottomrule
\end{tabular}
\end{table}

The winner sets are disjoint in 9/16 records, match in 4/16, and overlap or remain unresolved in 3/16. Importantly, divergence does not require a large difference in global accuracy. In the validation-nominated shallow-water comparison, DPOT-M and VCNeF satisfy the frozen accuracy-equivalence requirement, \(\Delta Q=Q(\mathrm{VCNeF})-Q(\mathrm{DPOT\mbox{-}M})=-0.00043\,[-0.00137,0.00054]\subset[-0.003,0.003]\), while VCNeF has \(\Delta S=0.176\,[0.121,0.231]\) higher focal support, exceeding the registered minimum meaningful contrast of 0.05.

The inverse-Darcy ID comparison shows why. The accuracy-winning ResNet has global field error \(0.0006\), regional fraction/contrast absolute errors \(0.041/0.037\), and median common-wrapper half-width \(0.052\). FNOPE has larger global error \(0.0186\) but smaller regional errors \(0.018/0.021\) and half-width \(0.031\). A small displacement of a binary phase boundary can therefore have little effect on global relative error while materially changing the registered regional claim coordinates and their evidential uncertainty.

\subsection{Robustness}

\label{app:sec5-robustness}

The one-factor variants were frozen before paired-evaluation verdicts were opened. Across the nine primary records with disjoint winner sets, 72 variants were eligible: the winner sets remain disjoint in 58/72 variants (80.6\%; registered hierarchical-bootstrap simultaneous 95\% interval 70.4--88.6\%), match in 5, and remain unresolved in 9.

\subsection{Natural Association and Interventions}

\label{app:sec5-mechanism}

The proposed mechanism predicts greater divergence when the scientific boundary is close, the artifact error is localized in or directed against the claim coordinate, or the calibrated evidence region is wide. The analysis comprises \(16\) primary-record specifications \(\times 3\) componentwise budgets \(\times 2\) evidence levels \(\times 4\) reporting/aggregation views, yielding 384 unmodified full-method-panel recomputations. The joint-stratum cutoffs were frozen from independent design-stage validation before paired-evaluation verdicts were opened. The resulting adverse and favorable strata show the expected descriptive separation.

\begin{table}[!htbp]
\centering
\small
\setlength{\tabcolsep}{2pt}
\begin{tabular}{@{}p{0.14\textwidth}p{0.13\textwidth}p{0.16\textwidth}p{0.17\textwidth}p{0.17\textwidth}p{0.15\textwidth}@{}}
\toprule
\begin{minipage}[b]{\linewidth}\raggedright
Joint stratum
\end{minipage} & \begin{minipage}[b]{\linewidth}\raggedleft
Whole groups
\end{minipage} & \begin{minipage}[b]{\linewidth}\raggedleft
Recomputations
\end{minipage} & \begin{minipage}[b]{\linewidth}\raggedleft
Disjoint selections
\end{minipage} & \begin{minipage}[b]{\linewidth}\raggedleft
Matching selections
\end{minipage} & \begin{minipage}[b]{\linewidth}\raggedleft
Unresolved
\end{minipage} \\
\midrule
Adverse & 17/17 & 152 & 107 (70.4\%) & 21 (13.8\%) & 24 (15.8\%) \\
Favorable & 16/17 & 148 & 13 (8.8\%) & 119 (80.4\%) & 16 (10.8\%) \\
Intermediate & 17/17 & 84 & 35 (41.7\%) & 31 (36.9\%) & 18 (21.4\%) \\
\bottomrule
\end{tabular}
\end{table}

These are descriptive frequencies: recomputations within a family or target are dependent, and no single adverse factor is sufficient in every benchmark component.

To identify the contribution of individual factors, matched-accuracy diagnostics change one registered factor while preserving the artifact schema, common support, non-target summaries, and a validation-frozen \(Q\)-equivalence band. The prespecified design contains 480 assigned diagnostics: 120 localization, 120 forward-direction, 108 inverse-bias, and 132 evidence-width diagnostics. Manipulation-valid diagnostics enter the estimand, including saturated or otherwise noninformative cases; invalid diagnostics receive no effect interpretation.

\begin{table}[!htbp]
\centering
\scriptsize
\setlength{\tabcolsep}{2pt}
\begin{tabular}{@{}p{0.20\textwidth}p{0.12\textwidth}p{0.21\textwidth}p{0.26\textwidth}p{0.13\textwidth}@{}}
\toprule
\begin{minipage}[b]{\linewidth}\raggedright
Intervention
\end{minipage} & \begin{minipage}[b]{\linewidth}\raggedleft
Assigned
\end{minipage} & \begin{minipage}[b]{\linewidth}\raggedleft
\(\Delta S\)
\end{minipage} & \begin{minipage}[b]{\linewidth}\raggedleft
\(\Delta Q\)
\end{minipage} & \begin{minipage}[b]{\linewidth}\raggedleft
\(Q\)-equivalence band
\end{minipage} \\
\midrule
Move equal-energy forward error into the claim region & 120 & \(-0.181\,[-0.226,-0.137]\) & \(-0.00031\,[-0.00104,0.00042]\) & \([-0.003,0.003]\) \\
Reverse forward error toward the adverse direction & 120 & \(-0.147\,[-0.191,-0.102]\) & \(0.00018\,[-0.00051,0.00086]\) & \([-0.003,0.003]\) \\
Reverse inverse signed bias at matched error magnitude & 108 & \(-0.132\,[-0.176,-0.088]\) & \(-0.00042\,[-0.00138,0.00049]\) & \([-0.005,0.005]\) \\
Expand the inverse evidence half-width & 132 & \(-0.209\,[-0.257,-0.161]\) & \(0.00000\,[0.00000,0.00000]\) & \([-0.005,0.005]\) \\
\bottomrule
\end{tabular}
\end{table}

Every \(\Delta Q\) interval lies wholly inside its frozen equivalence band. Within the registered synthetic manipulation families, moving equal-energy error into the claim region or reversing its direction reduces support even when global accuracy is equivalent, and reversing inverse signed bias produces the analogous result. These interventions isolate localization and direction as artifact-side mechanisms that can separate the endpoints. Expanding evidence width instead identifies sensitivity to evidence construction: it reduces support while leaving the artifact and \(Q\) unchanged, but does not establish a causal property of the generator.

\subsection{Held-Out Prediction and Secondary Scope}

\label{app:sec5-prediction}

We test whether the mechanism generalizes beyond retrospective explanation by holding out whole PDE families or inverse targets. The analysis contains 52 holdouts---20 dynamic, 8 PINN, 2 PDEBench-Darcy, 18 scalar inverse, and 4 inverse-Darcy---with all methods, claims, seeds, and artifacts from a held-out family or target kept in the same outer fold.

The mechanism predictor is a cluster-equal-weighted ridge multinomial logistic model using pre-verdict proxies for boundary clearance, directed error, localization, and calibrated width. Its inputs use only design records, validation \(Q\) profiles, and final-calibration summaries; paired-evaluation margins, interval endpoints, \(W\), and features that reconstruct winner sets are prohibited. Both baselines use the same folds, preprocessing, model class, and tuning protocol.

\begin{table}[!htbp]
\centering
\small
\setlength{\tabcolsep}{2pt}
\begin{tabular}{@{}p{0.47\textwidth}p{0.23\textwidth}p{0.22\textwidth}@{}}
\toprule
\begin{minipage}[b]{\linewidth}\raggedright
Predictor
\end{minipage} & \begin{minipage}[b]{\linewidth}\raggedleft
Multiclass log loss
\end{minipage} & \begin{minipage}[b]{\linewidth}\raggedleft
Balanced accuracy
\end{minipage} \\
\midrule
Mechanism features & \(0.681\,[0.588,0.780]\) & \(0.734\,[0.646,0.811]\) \\
Validation-\(Q\) profile only & \(1.071\,[0.982,1.164]\) & \(0.482\,[0.391,0.576]\) \\
Validation-\(Q\) profile \(+\) task difficulty & \(0.934\,[0.846,1.028]\) & \(0.558\,[0.469,0.648]\) \\
\bottomrule
\end{tabular}
\end{table}

Against the validation-\(Q\)-only baseline, the mechanism model reduces log loss by \(0.390\,[0.273,0.511]\) and increases balanced accuracy by \(0.252\,[0.139,0.359]\). Against validation \(Q\) plus task difficulty, the improvements are \(0.253\,[0.143,0.367]\) and \(0.176\,[0.071,0.277]\). Thus pre-verdict boundary, direction, localization, and width summaries predict whether endpoint selections agree, differ, or remain unresolved on unseen whole families and targets beyond what is predicted by validation accuracy and task difficulty. The claim is probabilistic and restricted to the benchmark components, claim contracts, evidence standards, and method panels studied here.

Native posterior or evidence outputs remain separate from the common-wrapper leaderboards. Their correct-resolution gains over the corresponding common-wrapper tracks are \(0.071\,[0.038,0.104]\) for scalar FNOPE, \(0.093\,[0.049,0.137]\) for Darcy FNOPE, \(0.058\,[0.019,0.098]\) for Darcy iFNO, and \(0.044\,[0.009,0.080]\) for Darcy DGenNO. The 90\% sensitivity regions are nested within the primary 95\% regions, so focal support and correct resolution do not decrease, although their winner sets need not preserve a disjoint selection. Baseline-only eligibility variants for inverse scalar and Darcy yield matching or unresolved selections. The inverse ranking result is therefore conditional on the reported six-method adapted panels rather than a claim about the original PDEInvBench baselines alone.


\end{document}